# Summarizing First-Person Videos
# from Third Persons' Points of Views

Hsuan-I Ho[1], Wei-Chen Chiu[2], and Yu-Chiang Frank Wang[1]

[1] Department of Electrical Engineering, National Taiwan University, Taiwan
{b01901029, ycwang}@ntu.edu.tw
[2] Department of Computer Science, National Chiao Tung University, Taiwan
walon@cs.nctu.edu.tw

**Abstract.** Video highlight or summarization is among interesting topics in computer vision, which benefits a variety of applications like viewing, searching, or storage. However, most existing studies rely on training data of third-person videos, which cannot easily generalize to highlight the first-person ones. With the goal of deriving an effective model to summarize first-person videos, we propose a novel deep neural network architecture for describing and discriminating vital spatiotemporal information across videos with different points of view. Our proposed model is realized in a semi-supervised setting, in which fully annotated third-person videos, unlabeled first-person videos, and a small number of annotated first-person ones are presented during training. In our experiments, qualitative and quantitative evaluations on both benchmarks and our collected first-person video datasets are presented.

**Keywords:** Video Summarization · First-Person Vision · Transfer Learning · Metric Learning

## 1 Introduction

Wearable and head-mounted cameras have changed the way how people record and browse videos. These devices enable users to capture life-logging videos without intentionally focus on particular subjects. Thus, the resulting first-person videos (or egocentric videos) would exhibit very unique content and properties when comparing to those of third-person ones. As pointed out by Molino *et al.* [18], the lack of sufficient structural information and repetitive content for first-person videos would limit viewing quality. Therefore, it is desirable to be able to highlight or summarize such videos for improving user viewing experiences.

With the goal of encapsulating informative segments from videos, video summarization aims at identifying the highlight video segments. Existing approaches for video summarization either select the most representative video segments [8,16] or detect particular or pre-defined visual structures or objects as summarized outputs [13,14,15]. With the recent development of deep learning, a recent work [28] successfully utilized deep neural networks for first-person video summarization, by using a pre-collected and annotated first-person video dataset.

Nevertheless, it is difficult to collect a large amount of fully annotated first-person video data (note that the dataset in [28] is not publicly available). To address such limitations, the technique of transfer learning becomes an alternative solution. To be more precise, it is possible for one to learn from annotated third-person videos and aim at transferring the learned model to summarize the first-person ones. However, since significant differences of visual appearance can be expected between third and first-person videos, how to apply and adapt existing third-person video summarization approaches would be a challenging task. Moreover, a satisfying first-person video summary should consist of segments important to both the recorder and the viewer. Without observing any annotated first-person video revealing such information, it would be difficult to learn an effective model for solving the corresponding task.

Existing transfer learning works have been focusing on alleviating the domain shift (or dataset bias) across data domains [19]. With the recent success of deep learning, architectures of deep neural networks have also been utilized to solve similar problems [21]. Recent deep learning based video summarization approaches like [11,17,30] did not explicitly address this issue. To advance transfer learning for first-person video summarization, one could utilize fully annotated third-person videos plus a number of annotated first-person videos for training their models. In order to increase the training set size of first-person videos without label information, one could further extend the above supervised domain adaptation setting to a more challenging yet practical semi-supervised one. That is, additional unlabeled first-person videos can be also presented during training. As a result, one not only requires to alleviate the domain (viewpoint) bias between first and third-person videos, how to learn deep neural networks in a semi-supervised setting needs to be also addressed.

In this paper, we propose a deep learning framework which performs cross-domain feature embedding and transfers highlight information across video domains. More specifically, our network architecture jointly performs domain adaptation (across third-and first-person videos) in a semi-supervised setting. That is, in addition to third-person videos with fully annotated highlight scores, first-person videos are also presented during training, while only a small portion of them are with ground-truth scores. Moreover, we further integrate a sequence-to-sequence model based on recurrent neural networks (RNN), which allows the exploitation of long-term temporal information for improved summarization.

In summary, our contributions are threefold: 1) By reducing the semantic gap between third and first-person videos, our proposed network transfers informative spatiotemporal features across video domains to perform first-person video summarization; 2) our network is able to handle unlabeled data during adaptation, which not only allows our model to be trained in a semi-supervised setting, possible overfitting due to a small amount of annotated first-person video data can be also alleviated; 3) in addition to the use of SumMe [7] dataset, we collect a larger-scale first-person video dataset for further evaluation, which is now available[3].

## 2    Related Works

**First-Person Video Summarization** Summarizing first-person videos has attracted the computer vision community in recent years [2,4,18]. Most existing approaches follow a basic workflow consisting of (1) visual feature extraction and (2) keyframe selection or scene segmentation, while the latter is typically subject to pre-defined criteria. For example, Lee *et al.* [13,14] select video frames containing important subjects and objects alongside visual diversity, while Bettadapura *et al.* [3] look for artistic properties in vacation-related videos. Lin *et al.* [15] train a context-specific highlight detector for each type of egocentric videos, which enables online summarization and solves the problem of data storage. Xu *et al.* [27] exploit gaze information to predict the attention given to video segments, resulting summarization that reflects the recorder preferences. However, the above methods are mainly applied to specific video contexts (e.g., daily live or cooking videos). Although a recent work by Yao *et al.* [28] learns an associated ranking function via deep metric learning to score video segments of 15 categories, it requires an enormous number of fully-annotated first-person videos (over 50 hrs) for training. This is why a semi-supervised transfer learning framework (as ours) for video summarization is practically preferable, with the goal of leveraging information across video domains for improved summarization.

**Deep Learning for Video Summarization** Some recent deep-learning based methods approach video summarization by solving a sequence-to-sequence problem, in which the video frames are encoded by Recurrent Neural Network (RNN) schemes. For example, Zhang *et al.* [30] propose a summarization model based on a bidirectional Long Short Term Memory (biLSTM) framework, which is trained on videos with annotated importance scores for keyframe selection. They additionally apply determinantal point process (DPP) to enhance the diversity of the chosen keyframes. Ji *et al.* [11] further extend such biLSTM models by integrating the attention mechanism. Their model considers temporal information in finer granularity when decoding the feature vectors of video segments generated by biLSTM.

Although supervised approaches exhibit promising video summarization results, existing datasets (with ground-truth data) for video summarization [7,23] are generally with smaller scales. For learning effective summarization models, it would be desirable to have a large number of labeled videos for training purposes. Several works thus attempt to utilize various techniques in order to address this issue. For example, Panda *et al.* [20] collect weakly annotated videos from YouTube 8M [1] and train their summarization model with auxiliary labels of activity classes; Sun *et al.* [24] train their highlight classifiers by utilizing a collection of YouTube videos that have been edited as positive training data, while the negative ones are retrieved from raw videos. Alternatively, Gygli *et al.* [9] present summarization models by collecting massive training pairs mined from GIF image websites. By advancing sequential generative adversarial networks, Mahasseni *et al.* [17] perform video summarization by predicting video keyframe distribution.

---

[3] The dataset and code is available on: https://github.com/azuxmioy/fpvsum

**Table 1.** Comparisons of existing video summarization datasets.

| Dataset | Type | Length | # of videos | Annotation/Score | Description |
|---|---|---|---|---|---|
| UT Ego [13] | 1st-person | 17 hr | 4 | Video frames which contain important people and objects | - Videos of daily activities in a uncontrolled setting |
| VideoSet [29] | 1st-person | >60 hr | 13 | Textual description for each 5-seconds video segment | - Provide textual labels for UT Ego [13] and DisneyWorld [6] <br> - **Not publicly available** |
| EgoSum+gaze [27] | 1st-person | >15 hr | 21 | 5 ~ 15 events selected by 5 camera wearers | - Daily lives videos together with gaze data <br> - **Not publicly available** |
| Yao *et al.* [28] | 1st-person | >100 hr | 600 | Fully annotated frame-level scores from 12 annotators | - 15 categories of GoPro sports videos mined from YouTube <br> - **Not publicly available** |
| SumMe [7] | 3rd-person | 50 min | 20 | Fully annotated frame-level scores from at least 15 annotators | - Raw user videos containing interesting events |
| TvSum [23] | 1st-person <br> 3rd-person | 14 min <br> 3 hr 30 min | 5 <br> 50 | Fully annotated frame-level scores from 20 annotators | - 50 YouTube videos in 10 categories from the TRECVid MED task |
| Proposed | 1st-person | 7 hr 56 min | 98 | Fully annotated frame-level scores from at least 10 annotators | - 14 categories of GoPro viewer-friendly videos selected from YouTube |

Nevertheless, the above approaches generally focus on summarizing third-person videos, or those with mixed type of videos [7,23] (i.e., no distinction between third and first-person ones). As noted above, highlighting first-person videos would be particularly challenging due to significant changes in visual content and appearances (plus, due to the lack of a sufficient amount of annotated training data). This is the reason why we choose to address first-person video summarization in a semi-supervised setting, and propose deep transfer learning techniques for solving this problem.

**Datasets for Video Summarization** Finally, we summarize the characteristics of existing datasets for both first-person and third-person video summarization in Table 1. UT Ego [13] annotates the keyframes including important objects and people in daily lives videos. VideoSet [29] provides extra textual labels for videos in UT Ego [13] and Disneyworld [6], including tools for summarization evaluation. EgoSum+gaze [27] consists of shot-level annotations obtained from camera wearers together with their gaze information. However, the context of the above first-person datasets are very limited (e.g., daily lives, cooking, etc. activities). Moreover, it is difficult for viewers to obtain frame-level importance scores due to their long duration and redundancy.

Yao *et al.* [28] first propose a large-scale dataset including frame-level annotations for various sports videos mined from YouTube. In contrast to two widely used dataset SumMe [7] and TvSum [23], most first-person videos mined from YouTube are either over-edited or overlong, which can result in very dif-ferent viewpoints across videos, plus frame discontinuity and annotation biases within the dataset. It is worth noting that, most first-person datasets for video summarization are not publicly available, this is the reason why one of our major contributions is to collect and release a first-person video dataset including viewer-friendly videos and unbiased importance scores for research purposes.

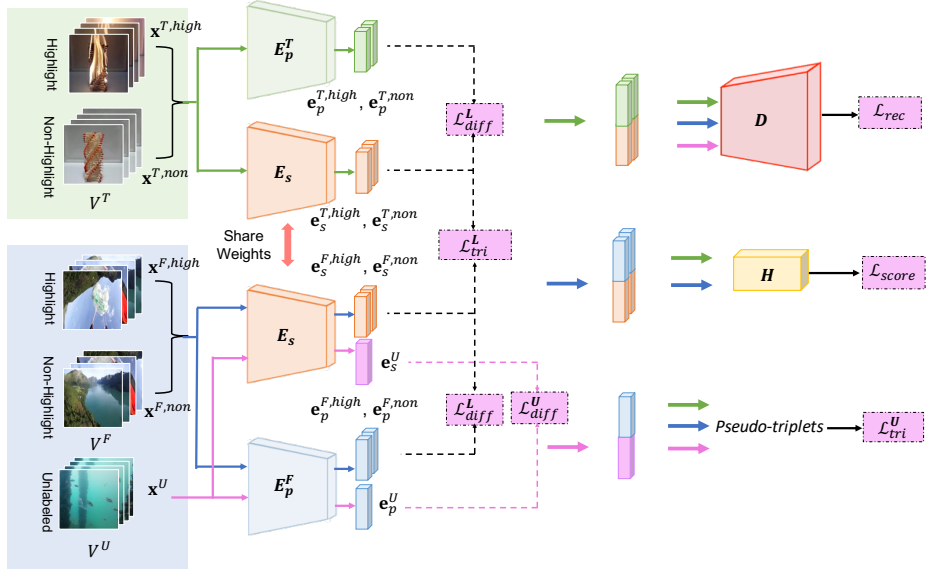

**Fig. 1.** Our first-person video summarization framework via semi-supervised domain adaptation. Note that fully annotated third-person videos $V^T$, unlabeled first-person videos $V^U$, plus a number of annotated first-person ones $V^F$ are presented during training. We have $\mathbf{x}$, $\mathbf{e}_s$, and $\mathbf{e}_p$ denote the input, *shared* and *private* features, respectively.

## 3  Proposed Framework

For the sake of completeness, we first explain the notations in this paper. We have an annotated video collection including a set of third-person videos $V^T = \left\{V_1^T, ..., V_M^T\right\}$ and few first-person videos $V^F = \left\{V_1^F, ..., V_N^F\right\}$, where $M$ and $N$ denote the numbers of third-and first-person videos respectively (where typically $M > N$). Their corresponding annotations (i.e., importance scores ) at the frame-level are $S^T = \left\{S_1^T, ..., S_M^T\right\}$ and $S^F = \left\{S_1^F, ..., S_N^F\right\}$. In addition, we have another set of first-person videos $V^U = \left\{V_1^U, ..., V_K^U\right\}$ without any annotation of importance scores, with the number of videos $K \gg N$. The goal of our work is to bridge the semantic gap across $V^T$, $V^F$ and $V^U$, so that the learned model can be applied for first-person video summarization.

The architecture of our proposed method is shown in Fig. 1, which consists of network components for cross-domain feature embedding and summarization.

The highlights of our method include: 1) our domain separation architecture learns shared and private features across video domains, while adapting cross-domain highlight information for summarization; 2) a self-learning scheme for leveraging unlabeled first-person video data, so that our model can be trained in a semi-supervised setting; and 3) our highlight detection network for exploiting long-term temporal information to improve final summarization. Details of our framework will be described in the following subsections.

### 3.1   Cross-Domain Feature Embedding

To adapt information across video domains for highlighting a particular video domain of interest, the first stage of our proposed network performs cross-domain feature embedding. More specifically, we aim at retrieving and transferring representative highlight information across third and first-person videos, while suppressing irrelevant features in each domain. This is achieved by performing cross-domain feature embedding via a domain separation structure. Inspired by [5], our network component for domain separation decomposes feature representations into of two subspaces: a *shared* subspace across video domains to extract domain-invariant information, and *private* subspaces which are unique to each domain for describing domain-specific properties.

Given video segments with fixed-length, i.e., $X^T = \left\{ \mathbf{x}_1^T, ..., \mathbf{x}_m^T \right\}$ and $X^F = \left\{ \mathbf{x}_1^F, ..., \mathbf{x}_n^F \right\}$ from $V^T$ and $V^F$, respectively, we view such segments as the basic elements in our framework for capturing video spatiotemporal information. The shared encoder $E_s$ in Fig. 1 encodes $\mathbf{x}^T, \mathbf{x}^F$ into domain-invariant representations $\mathbf{e}_s^T, \mathbf{e}_s^F$, while the private encoders $E_p^T$ and $E_p^F$ embed them into domain-specific features $\mathbf{e}_p^T$ and $\mathbf{e}_p^F$. These encoders are jointly learned with a decoder $D$ and two explicit loss functions: the reconstruction loss $\mathcal{L}_{rec}^L$ and difference loss $\mathcal{L}_{diff}^L$. Note that, $\mathcal{L}_{rec}^L$ encourages the reconstruction of $\mathbf{x}$ by decoder $D$ from the features concatenating shared and private representations $\mathbf{e} = \mathrm{concat}(\mathbf{e}_s, \mathbf{e}_p)$, which can be written as:

$$\mathcal{L}_{rec}^L = \sum_{i \in \{F,T\}} \left\| f_D(\mathbf{e}^i) - \mathbf{x}^i \right\|_2^2. \tag{1}$$

As for the difference loss $\mathcal{L}_{diff}^L$, it is imposed on the orthogonality between $\mathbf{e}_s$ and $\mathbf{e}_p$, and thus enforces $E_s$ and $E_p$ to capture different aspects of information (shared and private ones) from $\mathbf{x}$. Thus, $\mathcal{L}_{diff}^L$ is defined as:

$$\mathcal{L}_{diff}^L = \left\| \mathbf{E}_p^{T\top} \mathbf{E}_s^T \right\|_F^2 + \left\| \mathbf{E}_p^{F\top} \mathbf{E}_s^F \right\|_F^2, \tag{2}$$

where $\mathbf{E}_p^T$ and $\mathbf{E}_s^T$ are the matrices consisted of private and shared features of the third-person video segments in a batch. Likewise, $\mathbf{E}_p^F$ and $\mathbf{E}_s^F$ denote the matrices for the corresponding first-person embedded features. Exploiting the above domain separation architecture allows us to mitigate domain differences via shared feature embedding across video domains, and meanwhile retains sufficient domain-specific characteristic in the private subspaces for each domain.

Despite the above use of domain separation components for suppressing feature differences across video domains, there is *no* guarantee the shared encoder $E_s$ captures the semantics of highlight information from input videos. Hence, we further advance deep metric learning with the triplet network [10,22] for improved feature embedding. This not only allows us to better describe videos over third and first-person views but also reflects and shares highlight information across video domains for later summarization purposes.

To achieve the above goal, we divide the cross-domain videos in shared features with ground-truth score annotation into highlight and non-highlight subsets: $\{\mathbf{e}_s^{T,high}, \mathbf{e}_s^{T,non}\}$ and $\{\mathbf{e}_s^{F,high}, \mathbf{e}_s^{F,non}\}$, where $\mathbf{e}_s^{T,high}$ and $\mathbf{e}_s^{T,non}$ relate to the highlight and non-highlight third-person video subsets, respectively. Similarly, we have first-person video subsets $\{\mathbf{e}_s^{F,high}, \mathbf{e}_s^{F,non}\}$. To jointly eliminate video domain differences and adapt highlight information across domains, the triplets are built from a set of features $\{\mathbf{e}_s^{T,high}, \mathbf{e}_s^{T,non}, \mathbf{e}_s^{F,high}, \mathbf{e}_s^{F,non}\}$ to include feature pairs extracted from within and across-domain video data. Take $\{\mathbf{e}_s^{T,high}, \mathbf{e}_s^{T,non}, \mathbf{e}_s^{F,high}\}$ as examples, the corresponding triplet loss is calculated as:

$$\mathcal{L}_{tri}^{L} = max\left\{0, \mathcal{M} - \mathcal{D}_{cos}(\mathbf{e}_s^{T,high}, \mathbf{e}_s^{T,non}) + \mathcal{D}_{cos}(\mathbf{e}_s^{F,high}, \mathbf{e}_s^{T,high})\right\},\quad(3)$$

where $\mathcal{D}_{cos}(\mathbf{e}, \mathbf{e}') = 1 - \frac{\mathbf{e} \cdot \mathbf{e}'}{\|\mathbf{e}\|_2 \|\mathbf{e}'\|_2}$ returns the distance between the embedding features, and $\mathcal{M}$ denotes the margin for metric learning. Note that such losses are calculated for all triplets from within and cross-video data.

### 3.2   Self-Learning with Unlabeled First-Person Videos

The feature embedding network described in Sect. 3.1 allows us to identify domain-invariant and domain-specific features across video domains. However, since the number of labeled first-person videos is generally much smaller than that of third-person ones, it would be desirable to further exploit unlabeled first-person ones, so that possible overfitting can be alleviated.

In our proposed network, we thus introduce a *self-learning* component for addressing this task. As illustrated in Fig. 1, unlabeled segments of first-person videos pass through the embedding network, resulting in feature vectors $\mathbf{e}_p^U, \mathbf{e}_s^U$. The reconstruction loss $\mathcal{L}_{rec}^U$ and difference loss $\mathcal{L}_{diff}^U$, which are identical to $\mathcal{L}_{rec}^L$ and $\mathcal{L}_{diff}^L$ respectively but applied for unlabeled $\mathbf{e}^U$ and $\mathbf{E}^U$, are also used in training:

$$\mathcal{L}_{rec}^U = \left\|f_D(\mathbf{e}^U) - \mathbf{x}^U\right\|_2^2, \mathcal{L}_{diff}^U = \left\|\mathbf{E}_p^{U\top}\mathbf{E}_s^U\right\|_F^2.\qquad(4)$$

To further utilize unlabeled first-person videos together with the annotated cross-domain data for improved cross-domain feature embedding, we extend the above learning scheme by generating pseudo triplets, which allows us to finetune the above cross-domain embedding network. To be more precise, in a subset built upon samples from $\{\mathbf{e}^T, \mathbf{e}^F\}$, the farthest and nearest features with respect to each reference $\mathbf{e}^U = \text{concat}(\mathbf{e}_s^U, \mathbf{e}_p^U)$ would be viewed as the negative pair $\mathbf{e}^-$ and positive pair $\mathbf{e}^+$ for $\mathbf{e}^U$, thus forming pseudo triplets. Therefore, without observing ground-truth annotation scores, the loss of such a pseudo triplet $\mathcal{L}_{tri}^{U}$ can be calculated for each unlabeled first-person video:

$$\mathcal{L}_{tri}^{U} = max\left\{0, \mathcal{M} - \mathcal{D}_{cos}(\mathbf{e}^{U}, \mathbf{e}^{-}) + \mathcal{D}_{cos}(\mathbf{e}^{U}, \mathbf{e}^{+})\right\}. \qquad (5)$$

With the above self-learning strategy, we are now able to jointly exploit both supervised and unlabeled video data during training, so that network components $E_p^F, E_s$ and $D$ can be updated accordingly.

### 3.3   From Segment to Sequence Based Highlight Detection

From the above subsections, we see that the first stage of our network performs feature embedding across video domains in a semi-supervised setting, while both representation and discriminative highlight information are preserved in the feature space. Since the focus of our network is to perform first-person video summarization, we finally introduce a highlight detection network. As depicted in Fig. 1, this would additionally enforce the resulting joint features of first-person videos to exhibit sufficient highlight information.

With our use of highlight and non-highlight scores, the introduced highlight detection network serves as a binary classifier, which distinguishes between the associated video segments accordingly. Thus, we do not consider ranking loss as [9,24,28] did. Instead, following [17], we apply classification loss for our highlight detection model.

To detail this highlight detection process, we have concatenated features $\{\mathbf{e}_1, ..., \mathbf{e}_B\}$ as inputs to the highlight detection network $H$ for predicting the importance scores $\hat{\mathbf{s}}_i = f_H(\mathbf{e}_i)$. Note that $\mathbf{e}_i = \mathrm{concat}(\mathbf{e}_{p,i}, \mathbf{e}_{s,i})$, and B as the number of instances in each batch. The scoring loss between the predicted and ground-truth scores is calculated as:

$$\mathcal{L}_{score} = -\frac{1}{B}\sum_{i=1}^{B}\mathbf{y}_i \cdot \log\left(\hat{\mathbf{s}}_i\right), \qquad (6)$$

where each ground-truth scores $s_i$ are converted to 2-D one-hot vectors $\mathbf{y}_i = (0, 1) \vee (1, 0)$, and the network $H$ returns a 2-D softmax prediction $\hat{\mathbf{s}}_i$ instead of an 1-D scalar $\hat{s}_i$.

To extend the above segment-based video highlight detection into sequence-based prediction, we further extend the proposed network architecture to a sequence level, i.e., the feature output of the embedding network is now served as the input to a recurrent neural network (RNN). As suggested in [11,17,30], integration with such RNN-based components allows one to observe long-term dependency between segments within a video sequence. While recent RNN-based models can be easily applied and integrated into our framework (e.g., LSTM-based models [30], adversarial LSTM networks [17], and attention-based encoder-decoder networks [11]), we particularly take the video summarization LSTM (vsLSTM) in [30] for our use. We note that vsLSTM consists of a bidirectional LSTM (biLSTM) cell followed by a single-hidden-layer MLP. The biLSTM cell takes a sequence of concatenated features $\mathcal{E} = \{\mathbf{e}_1, ..., \mathbf{e}_t\}$ as inputs and returns both forward hidden states $\mathbf{h}^{forward} = \{\overrightarrow{\mathbf{h}}_1, ..., \overrightarrow{\mathbf{h}}_t\}$ and backward hidden states $\mathbf{h}^{backward} = \{\overleftarrow{\mathbf{h}}_1, ..., \overleftarrow{\mathbf{h}}_t\}$. These observed hidden states would exploit and preserve semantic information across time periods. Upon the introduction of LSTM-based models, a single-hidden-layer MLP can be directly deployed for predicting the importance scores $\hat{\mathbf{s}} = \{\hat{\mathbf{s}}_1, ..., \hat{\mathbf{s}}_t\}$ with inputs $\mathbf{h}^{forward}$, $\mathbf{h}^{backward}$, and $\mathcal{E}$. As a result, the scoring loss for updating vsLSTM can be calculated via (6).

### 3.4  Learning of Our Network

It is worth noting that, our proposed network allows end-to-end training, which updates the parameters for each component by calculating the following loss:

$$\mathcal{L}_{total} = \mathcal{L}_{tri} + \alpha \cdot \mathcal{L}_{rec} + \beta \cdot \mathcal{L}_{diff} + \gamma \cdot \mathcal{L}_{score}, \tag{7}$$

where $\alpha, \beta, \gamma$ are hyperparameters that control the interaction of overall loss. Except for $\mathcal{L}_{score}$ which relies on video data with ground-truth scores, the remaining losses are calculated and summed over labeled and unlabeled data.

As for the learning of sequence-based highlight detection network, a two-stage training scheme is implemented. That is, we first train the feature embedding network using video segment pairs as shown in Fig. 1 (i.e., segment-based highlight detection network), followed by jointly training of RNN and the resulting network using consecutive video segments as inputs.

## 4  Experiments

### 4.1  Datasets

We now describe the datasets (including the one we collect) for experiments. Two publicly available datasets with full annotations, **SumMe** [7] and **TV-Sum** [23], are recently used to evaluate the performance of video summarization task. Both cover a variety of video contexts for summarization purposes. **SumMe** consists of 25 user videos with a length varying from 1 to 6 minutes, in which the annotations of frame-level importance scores are provided. Within this dataset, there are five first-person videos, "Base jumping, Bike Polo, Scuba, Valparaiso Downhill, Uncut Evening Flight", which are applied as test data for quantitative evaluation and comparisons. On the other hand, **TVSum** consists of 50 third-person videos collected from YouTube, and each of them is annotated with frame-level importance scores. The videos in this dataset are viewed as the third-person labeled data for our training purposes.

Nevertheless, the number of first-person videos in SumMe is far from sufficient for training effective deep learning models for summarization. Thus, following the procedure of [28], we additionally create a new first-person dataset containing various categories of first-person videos from YouTube along with corresponding frame-level importance scores.

**Table 2.** Descriptions and properties of video datasets considered in the experiments

| Dataset | Video type | Total length | # of videos | Usage | Annotations |
|---|---|---|---|---|---|
| SumMe | 1st-person | 14 min | 5 | Testing | Yes |
| | 3rd-person | 50 min | 20 | Training ($V^T$) | Yes |
| TvSum | 3rd-person | 210 min | 50 | Training ($V^T$) | Yes |
| FPVSum | 1st-person labeled | 162 min | 56 | 25% Training ($V^F$) 55% Training ($V^U$) 20% Testing | Yes |
| | 1st-person unlabeled | 314 min | 42 | Training ($V^U$) | None |

During the collection of the first-person videos, we found that a large number of such videos on YouTube are not raw videos but edited ones, consisting of obvious frame discontinuity (selected/edited by users), transitions of point-of-view, and unrelated contents. Thus, they cannot be directly applied and added to the data collection for training/testing. Another observation is about the annotation collection. We observe that most annotators would lose concentration on assigning scores for long videos. Therefore, we collect a first-person video summarization dataset **FPVSum** with a total number of 98 videos. Excluding the edited or discontinuous videos, our collected video data are from 14 categories with varying lengths (over 7 hours in total). For each category, about 35% of the video sequences are annotated with ground-truth scores by at least 10 users, while the remaining are viewed as the unlabeled ones.

Complete discussions and comparisons of the datasets considered are listed in Table 2. We note that, when evaluating the performance of first-person video summarization, only our proposed model can be realized in a semi-supervised setting, while existing baseline and state-of-the-art approaches cannot handle unlabeled first-person videos during training.

### 4.2   Experimental Setup

**Evaluation Metrics** We follow the criteria used in [17,30] for video summarization evaluation, with the length of video summaries less than 15% of the total length of the original video. Let $\mathcal{A}$ be the set of generated summaries, and $\mathcal{B}$ is the set of video segments selected by user-annotated importance scores, the resulting precision $\mathcal{P}$ and recall $\mathcal{R}$ are defined as:

$$\mathcal{P} = \frac{\text{total overlap duration of } \mathcal{A} \text{ and } \mathcal{B}}{\text{total duration of } \mathcal{A}}, \tag{8}$$

$$\mathcal{R} = \frac{\text{total overlap duration of } \mathcal{A} \text{ and } \mathcal{B}}{\text{total duration of } \mathcal{B}}. \tag{9}$$

Thus, the F-measure is computed as $\mathcal{F} = 2 \times \mathcal{P} \times \mathcal{R}/(\mathcal{P} + \mathcal{R}) \times 100\%$. In addition, we further calculate the area-under-curve (AUC) values based on the resulting precision-recall curves, which allow us to perform detailed comparisons with respect to different lengths of summaries.

**Methods of Interest** We compare our work with four baselines (noted as **Random**, **Uniform**, **DSN** [5] and **C3D** [26]) and two state-of-the-art supervised video summarization models: **TDCNN** [28] and **vsLSTM** [30]. We first describe sequentially how the four baselines are obtained.

- **Random**: 15% of segments from each test video are randomly sampled as the highlight.
- **Uniform**: Instead of random sampling, 15% of segments from each test video are equidistantly selected as the highlight.
- **DSN**: The direct use of domain separation networks (DSN) [5] for performing cross-view video summarization.
- **C3D**: We extract C3D [26] pre-trained features for each video segment. A highlight classifier taking C3D features as the inputs is trained by the classification loss (6). The segments with top 15% prediction scores in each video are selected as the highlight outputs.

As for the two state-of-the-art methods, their original objectives and experimental settings are different from our semi-supervised one. Thus, we cannot directly report and compare their performances. Instead, we implement their works with the following settings for fair comparisons.

- **TDCNN**: Although [28] originally designs a two-stream network that exploits two visual features (i.e., AlexNet [12] and C3D) in their model, we compare TDCNN (C3D) only in our experiments for the sake of fairness. We train a temporal highlight detection network which is built upon a 6-layers fully connected Siamese network and outputs importance scores. The loss function for the TDCNN classifier is defined as $\mathcal{L}_{pair} = max\left\{0, 1 - s(\mathbf{x}^{high}) + s(\mathbf{x}^{non})\right\}$, where $s(\mathbf{x}^{high})$ and $s(\mathbf{x}^{non})$ are the scores of highlight and non-highlight segments. The positive and negative pairs of training data for learning the Siamese network of TDCNN classifier is produced by following the same criteria described in Sect. 3.1.
- **vsLSTM**: As shown in [30], it is implemented as an architecture of stacking a video feature extractor, a biLSTM with 256 hidden units, and a single-hidden-layer MLP, where the parameters of vsLSTM are learned by using the mean squared loss. Note that the original GoogLeNet [25] feature extractor is now replaced by the same C3D architecture. We further experiments another variant of vsLSTM using classification loss as (6), which is utilized in [17].

Both TDCNN and vsLSTM models are trained with $V^T$ and $V^F$, and the resulting summaries are selected from segments with top 15% scores. In addition, we implement another two variants of our approach for controlled experiments:

- **Ours w/o** $V^U$: We train our model in Fig. 1 with supervised training data $V^T$ and $V^F$ only, i.e., a fully-supervised version without observing unlabeled first-person videos.
- **Ours (non-sequential)**: Instead of RNN, we apply a 2-layer fully connected network as the highlight detection network in Fig. 1. Note that this variant is applicable to both fully-supervised and semi-supervised settings.

**Table 3.** Performance evaluation and comparisons on first-person video summarization in terms of F-measures and AUC values. Note that only Ours utilizes unlabeled first-person videos during training, while vsLSTM+ replaces its original MSE loss by our classification loss.

| Method | | F-measure | | AUC value | |
|---|---|---|---|---|---|
| | | **SumMe** | **FPVSum** | **SumMe** | **FPVSum** |
| Baseline | Random | 16.312 | 15.071 | — | — |
| | Uniform | 15.053 | 15.670 | — | — |
| | DSN [5] | 22.658 | 19.345 | 0.2075 | 0.1662 |
| | C3D [26] | 26.945 | 19.595 | 0.2091 | 0.1938 |
| Non-sequential | TDCNN [28] | 28.623 | 31.174 | 0.2340 | 0.2658 |
| | Ours w/o $V^U$ | 35.272 | 37.098 | 0.2489 | 0.2904 |
| | **Ours** | **38.649** | **38.409** | **0.2733** | **0.2962** |
| Seuqential | vsLSTM [30] | 29.850 | 19.901 | — | — |
| | vsLSTM+ | 31.468 | 26.204 | 0.2421 | 0.2266 |
| | Ours w/o $V^U$ | 35.980 | 37.366 | — | — |
| | **Ours** | **41.991** | **38.572** | **0.3165** | **0.3120** |

### 4.3 Quantitative Evaluation

**Comparisons** Table 3 summarizes the quantitative results of our framework, baselines, and the state-of-the-art video summarization algorithms. When comparing non-sequential based methods (including baselines), our model produced favorable results due to learning of cross-domain feature embedding and exploitation of unlabeled training data $V^U$. Recent approaches of C3D and TDCNN were not able to achieve comparable results due to their lack of ability in learning from data across video domains. We also observe that the direct use of DSN to summarize videos across data domains without exploiting any cross-domain label information could not yield satisfactory performances.

We note that the use of our proposed model without observing unlabeled first-person video training data (i.e., **Ours w/o** $V^U$ in Table 3) still performed against the above state-of-the-art methods. This again verifies the effectiveness of our cross-domain embedding framework. As shown in Table. 3, the full version of our method achieved the best performance, which was above those reported by recent recurrent models of vsLSTMs. This is due to the fact that vsLSTM only takes C3D features as inputs, and is not designed to deal with cross-domain video data.

**Analysis of our Network Design** We now perform controlled experiments using variants of our model in the non-sequential setting. As noted in Table 4, Ours$^\dagger$ denotes our model using only shared encoder to describe cross-domain data, and Ours$_{ranking}$/Ours$_{MSE}$ are the ones using the associated losses, which were suggested in [9,28,30]. Finally, Ours* represents our model without enforcing the pseudo triplet loss in (5), while Ours** indicates the supervised version of our model (i.e., without observing any unlabeled first-person videos). From the results listed in Table 4, it is clear that our full model achieved the best performance performed. Thus, our model design and integration of the above components are desirable for cross-domain video summarization.

**Table 4.** Analysis of our network design and settings. Note that Ours$^\dagger$ indicates our model without using any private encoders, Ours$_{\mathrm{ranking}}$/Ours$_{\mathrm{MSE}}$ denote the uses of ranking and MSE losses as the scoring loss in our model, Ours* is the version excluding the pseudo triplet loss $\mathcal{L}^U_{tri}$, and Ours** represents the version without observing unlabeled data $V^U$.

| Method | Cross-domain embedding | | Scoring losses | | Unlabeled videos | | |
|--------|------------|--------|------------------------|------------------|--------|--------|--------|
| | TDCNN [28] | Ours$^\dagger$ | Ours$_{\mathrm{ranking}}$ | Ours$_{\mathrm{MSE}}$ | Ours* | Ours** | Ours |
| **SumMe** | 28.623 | 29.754 | 28.435 | 29.078 | 34.252 | 35.272 | **38.649** |
| **FPVSum** | 31.174 | 31.020 | 34.007 | 36.046 | 35.485 | 37.098 | **38.409** |

### 4.4 Example Visualization Results

Fig. 2 shows example summarization results of a challenging first-person test video "Valparaiso Downhill" in SumMe dataset (with ground-truth scores provided). This video is 3-minute long recorded by a camera mounted on a helmet. It is a typical first-person video since the video content reflects the motion of the recorder (i.e., the bike rider) and no specific objects are intentionally focused. The blue bars in Fig. 2 indicate frame-level ground-truth scores annotated by users. The interval colored in green, red and yellow correspond to summaries generated by our model, vsLSTM, and TDCNN, respectively. The red horizontal lines represent the threshold for splitting highlight and non-highlight parts, as described in Sect. 3.1.

It is worth noting that, there are two particular actions, "jumping" and "going downstairs", in this video. These two action types are very unique and closely related to first-person videos, and thus are generally not present in third-person videos like those in SumMe and TvSum dataset. We see that, while both TDCNN and vsLSTM failed to predict such highlight moments, our model was able to produce satisfactory summarization outputs due to the exploitation of information across annotated third and first-person videos (including unlabeled ones).

## 5 Conclusions

In this paper, we proposed a novel deep learning architecture of first-person video summarization. Our network uniquely integrates modules of cross-domain feature embedding and highlight detection in a recurrent framework, which allows the extraction and adaptation of spatiotemporal discriminative highlight information across video domains. Moreover, to alleviate possible overfitting due to a small amount of labeled first-person video data during training, the introduced self-learning scheme further allows us to exploit information observed from unlabeled first-person videos (i.e., a semi-supervised setting). In addition to evaluation of benchmark datasets, we additionally collect a new first-person video dataset. Quantitative and qualitative experimental results confirmed the effectiveness and robustness of our proposed model for first-person video summarization.

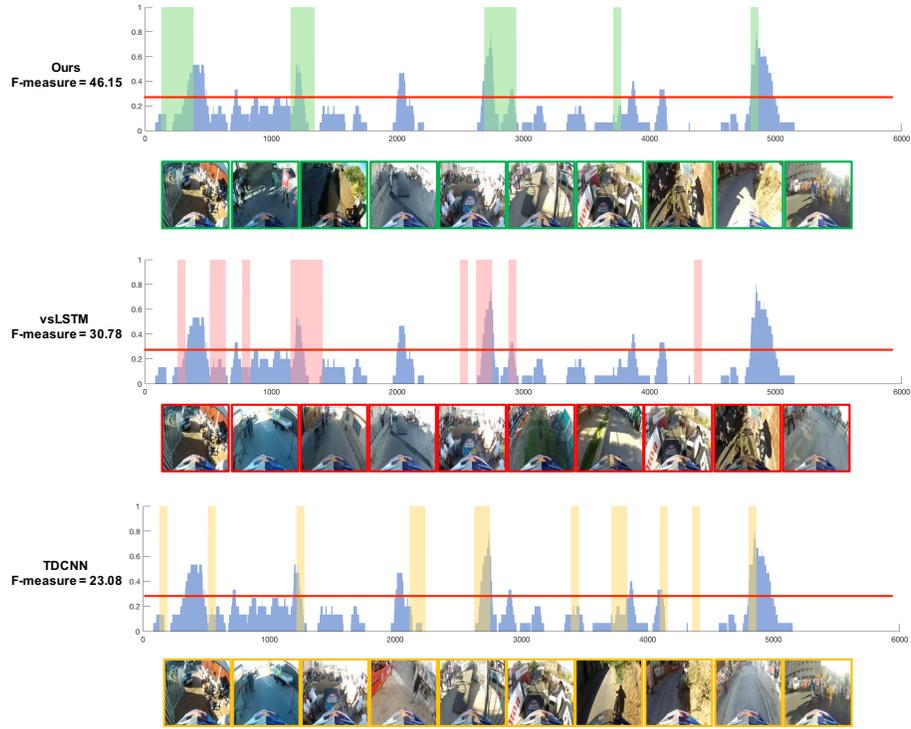

**Fig. 2.** Example summarization results of video "Valparaiso Downhill" from SumMe. The ground-truth annotation scores are shown in blue, while the predicted summaries from our model, vsLSTM, and TDCNN are shown in green, red and yellow, respectively. Note that the red horizontal line indicates the threshold which splits the scores into highlight and non-highlight ones. Note that our method produces desirable results by detecting movements such as "jumping" and "going downstairs". In contrast, others fail to capture the moments with a large number of false predictions.

**Acknowledgments** This work was supported in part by the Ministry of Science and Technology of Taiwan under grants MOST 107-2636-E-009-001 and 107-2634-F-002-010.

# Supplementary Material for
# "First-Person Video Summarization
# From Third Person's Point of Views"

Hsuan-I Ho[1], Wei-Chen Chiu[2], and Yu-Chiang Frank Wang[1]

[1] Department of Electrical Engineering, National Taiwan University, Taiwan
{b01901029, ycwang}@ntu.edu.tw
[2] Department of Computer Science, National Chiao Tung University, Taiwan
walon@cs.nctu.edu.tw

## 1 Dataset

**First-Person Video Data** We provide the details of our proposed first-person video dataset, FPVSum. This dataset is collected from YouTube by following the procedure of [7]. That is, we select 10 video categories from [1, 5, 7] plus 4 new ones (as listed in Table A). When collecting this video dataset, we focus on continuous first-person videos only (i.e., no transition within or between points of views); moreover, videos with unrelated contents will be excluded. Therefore, a total number of 98 first-person videos are obtained. Table A lists the videos of 14 categories. As discussed later, we will explain how the annotation is provided for selected videos for training, test, and evaluation purposes.

Table A: Descriptions and properties of our proposed FPVSum dataset. Note that (a) denotes the total length, (b) lists the numbers of highligh/non-highlight segments, and (c) shows the number of annotated/total number of frames.

| Category | (a) | (b) | (c) | Cronb. $\alpha$ | f-measure |
|---|---|---|---|---|---|
| Biking | 38m 22s | 51 / 290 | 20595 / 67669 | 0.879 | 0.414 |
| Bikepolo | 32m 31s | 40 / 323 | 23729 / 54270 | 0.733 | 0.252 |
| Boxing | 45m 39s | 72 / 347 | 25312 / 77237 | 0.754 | 0.294 |
| HorseRiding | 54m 39s | 48 / 307 | 21491 / 98369 | 0.954 | 0.609 |
| Jumping | 22m 25s | 43 / 208 | 15230 / 39279 | 0.875 | 0.422 |
| LongBoarding | 28m 32s | 58 / 300 | 21636 / 49335 | 0.771 | 0.297 |
| Motor | 24m 24s | 41 / 232 | 16545 / 39337 | 0.907 | 0.466 |
| Parkour | 21m 49s | 41 / 232 | 16561 / 35411 | 0.838 | 0.337 |
| Plane | 29m 50s | 61 / 271 | 20069 / 53787 | 0.753 | 0.279 |
| RockClimbing | 49m 17s | 80 / 377 | 27565 / 88709 | 0.495 | 0.244 |
| Scuba | 44m 28s | 98 / 412 | 30773 / 80089 | 0.618 | 0.225 |
| Skate | 23m 46s | 15 / 153 | 10263 / 40733 | 0.890 | 0.457 |
| Ski | 38m 3s | 66 / 269 | 20250 / 63522 | 0.870 | 0.431 |
| Surfing | 22m 16s | 48 / 158 | 12581 / 40098 | 0.903 | 0.524 |
| Total | 476m 1s | 762 / 3879 | 282600 / 827845 | 0.783 | 0.362 |

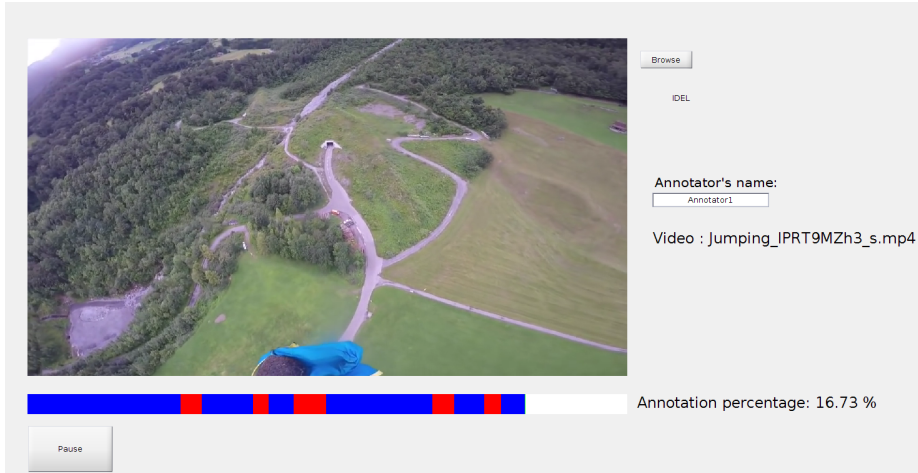

Fig. A: Our human interface for highlight annotations. For a given input video, the blue and red color bars denote non-highlight and highlight segments selected by a user, respectively.

**Annotations** We follow [1] to perform video annotation. That is, given each video, annotators are asked to produce a summary that contains most of its important content and highlight segments using our designed human interface shown in Fig. A. The interface shows each video excluding its audio track, ensuring annotators select highlight based on visual content only. Annotators are able to use the interface for moving forward and backward and modify their annotations at any time. The details of our annotation process are shown as follows:

- The annotators require to select highlight/non-highlight segments in each video. They need to finish watching each video once, then they start the labeling process.
- The annotators are asked to select the video parts which they consider interesting or important (i.e., mark the parts to red color using the interface in Fig. A). We note that an interesting part being marked may vary in any length.
- The annotators are encouraged to produce the summary which accounts for 10% to 20% of the full video length.
- Each frame would get an importance score which indicates how many annotators mark on this frame. We finally select frames ranked in the top 15% of all video frames as the highlight ones.

The consistency of human annotation for our FPVSum dataset can be evaluated by two metrics, Cronbach $\alpha$ and pairwise f-measure, which are both utilized

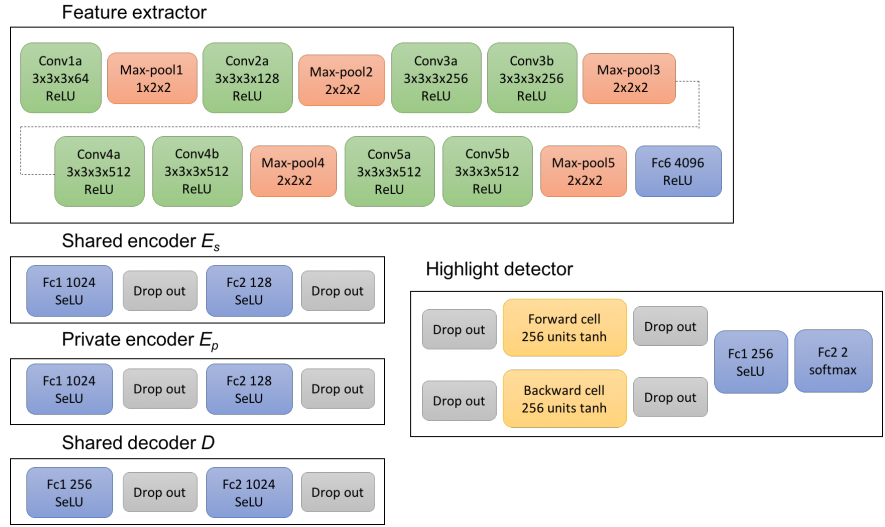

Fig. B: Our network topology for the components of feature extractor, feature embedding, and highlight detector.

to evaluate that of SumMe in [1]. Table A shows both human consistency metrics in each category.

As noted in our manuscript, videos with long durations typically result in inconsistent annotation scores from the users, since he/she tends to lose concentration when viewing/assigning highlight or non-highlight labels. Thus, the collected video sequences have durations of about 1 to 6 minutes for each particular category. Finally, about 65% of the videos are regarded as unlabeled data for learning. As for the remaining ones with ground truth scores (which are annotated by 10 different users), about 80% will be randomly chosen for training and the rest for testing.

## 2   Implementation Details

**Inputs**  In our first-person video summarization framework, we take video segments with fixed-length as the basic elements for capturing spatiotemporal information in the video. In particular, a video is split into a series of 2-second segments, and each segment is composed of 16 video frames (i.e., videos are down-sampled to 8 fps). We further categorize all video segments into highlight and non-highlight subsets according to their importance scores (segments of the top 15% importance scores are highlight ones, while the rest are non-highlight ones). The total number of highlight/non-highlight segments in FPVSum is shown in Table. A. Together with the videos from other datasets (i.e, SumMe and TvSum as listed in Table 2 of the main submission), we generate extensive training sets within and across first- and third-person highlight/non-highlight subsets.

**Network Structures** We first train a feature extractor for capturing spatiotemporal information in each video segment. We adopt the architecture of 3-Dimensional Convolutional Networks (i.e., C3D [6]) as our feature extractor, in which its weights are initialized by the C3D learned from Sport1M [3] video classification dataset while further fine-tuned in our training procedure of video summarization. The feature (4096-d) yielded from the fc-6 layer of extractor serves as the input of the cross-domain feature embedding network.

Our cross-domain feature embedding network consists of two private encoders, a shared encoder, and a shared decoder. Each encoder is a two-layer fully connected network (1024, 128 SeLU units), and the shared decoder has a two-layer fully connected structure (256, 1024 SeLU units). The sequential highlight detection network consists of a biLSTM with 256 hidden units in the both forward and backward cells followed by a 256-units fully connected layer and a softmax output layer. We present detailed network topology in Fig. B.

**Parameter Settings** To train the proposed model, we perform a two-stage optimization process as we mentioned in the main paper. That is, we first train the feature embedding network with segment-based highlight detector, where each segment is treated independently, then perform joint training of sequential highlight detector by using sets of consecutive video segments as input.

To be detailed, in the first stage we train our feature embedding network based on the 400K training sets generated from both first- and third-person highlight/non-highlight subsets. The margin parameter $\mathcal{M}$ of triplet loss is set as 1.2 and the size of shared and private features is 128. We train our network using Adam optimizer with a batch size of 8, first- and second-momentum of 0.9 and 0.99, and dropout probability of 0.8. We use the hyperparameters $\alpha = 0.5, \beta = 10^3, \gamma = 1.0$ to balance overall losses. The learning rate of the feature embedding network is set to $10^{-4}$ while the segment-based highlight detector is set to $10^{-5}$. The network is trained in total 50K steps. We note that, since the unlabeled data needs pseudo labels as described in Section 3.2 of our main paper, they are used after 10K steps of training.

In the second stage, consecutive video segments are used for jointly learning the parameters of the sequential highlight detector. We optimize our network by Adam optimizer with a batch size of 4, first- and second-momentum of 0.9 and 0.99, dropout probability of 0.8. The learning rate of the sequential highlight detector is set as $10^{-5}$ whereas the feature embedding network is finetuned with a learning rate of $10^{-6}$. The overall network is trained in total 3K steps.

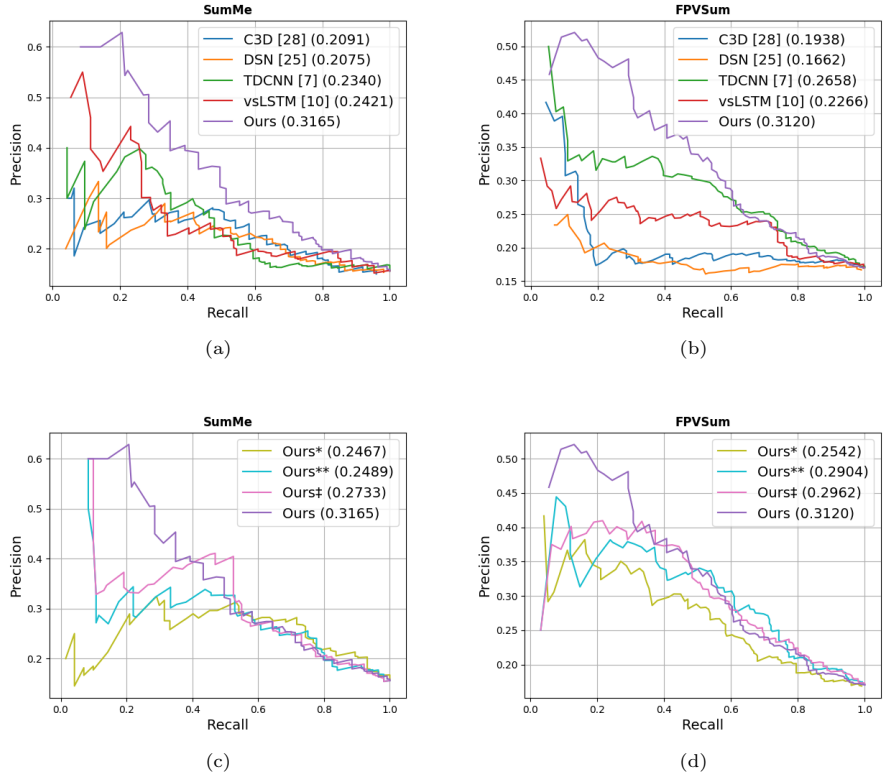

(a)

(b)

(c)

(d)

Fig. C: Precision-recall curves and AUC scores for SumMe and FPVSum. Note that (a) and (b) compare the P-R curves of several recent approaches, while (c) and (d) evaluate those of our variants (i.e., those listed in Table 4). The number followed by each method/model indicates the AUC score (e.g., 0.3165 for Ours on SumMe).

## 3 Additional Results

### 3.1 Precision-Recall Curves

In the main article, we follow the settings and evaluation metrics as those in [1, 2, 4, 5, 8], and compare f-measures of different methods. We additionally consider precision-recall curves and the corresponding area-under-curve (AUC) values for further evaluation.

Figures C(a) and (b) present the P-R curves and AUC scores of different methods on both SumMe and FPVSum datasets. It can be seen that our proposed model consistently performed against recent deep learning methods. On the other hand, Figures C(c) and (d) compare P-R curves and AUC scores of different variants of our model (i.e., those presented in Table 4). Note that **Ours**‡ indicates the non-sequential version of our model which use only fully connected

layers instead of RNN as the final highlight classifier. With such ablation studies, we again verify the contributions of the introduced components, which support the full version of our model for cross-domain video summarization.

### 3.2 Visualization

In this section, we show additional visualization results of testing videos. As in the main paper, the user-annotated scores (ground truth) are shown in blue, while the predicted summaries from our works, vsLSTM, and TDCNN are shown in green, red and yellow, respectively. The red horizontal line split the scores into highlight (i.e., top 15%) and non-highlight ones.

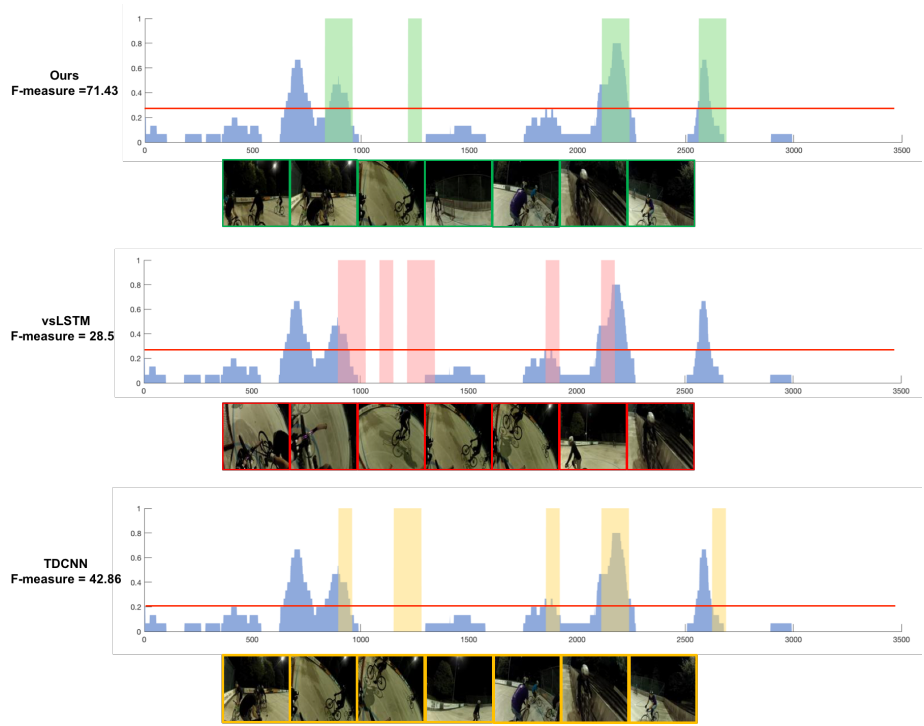

Fig. D: Performance comparisons on the video "Bike Polo" from SumMe. Note that the predicted highlight segments are denoted in green, red, and yellow for the methods of ours, vsLSTM, and TDCNN, respectively. We see that our summarization result is able to capture three precise highlight moments (e.g., shoot, goal, etc.) whereas others contained only parts of them.

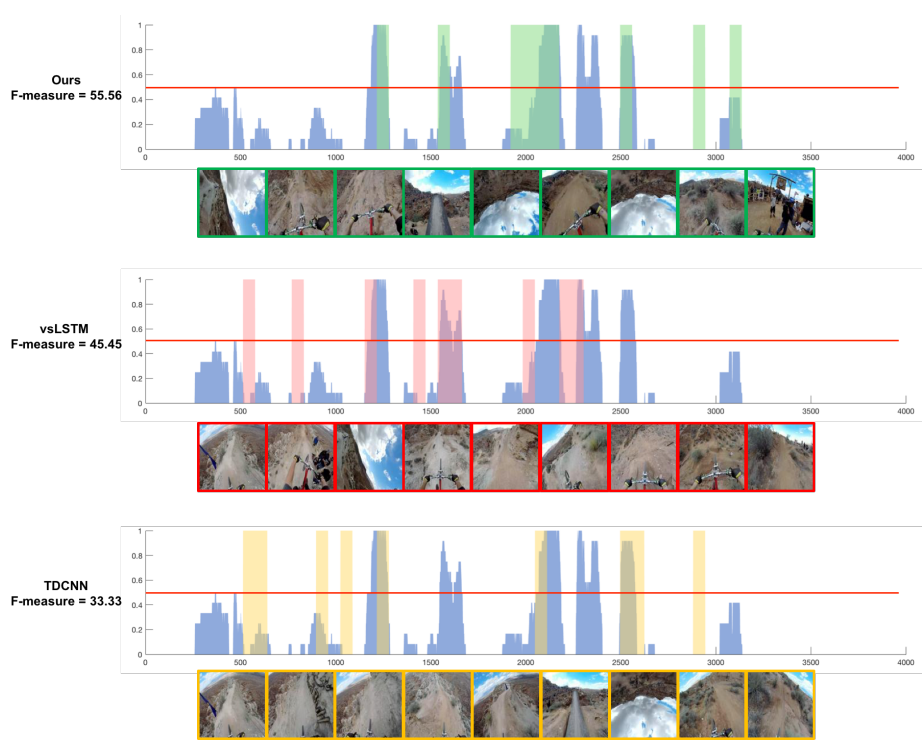

Fig. E: Performance comparisons on the first-person mountain biking video from FPV-Sum. We note that our summarization result includes unique moments in first-person biking videos such as "360° backflip" and "landing".

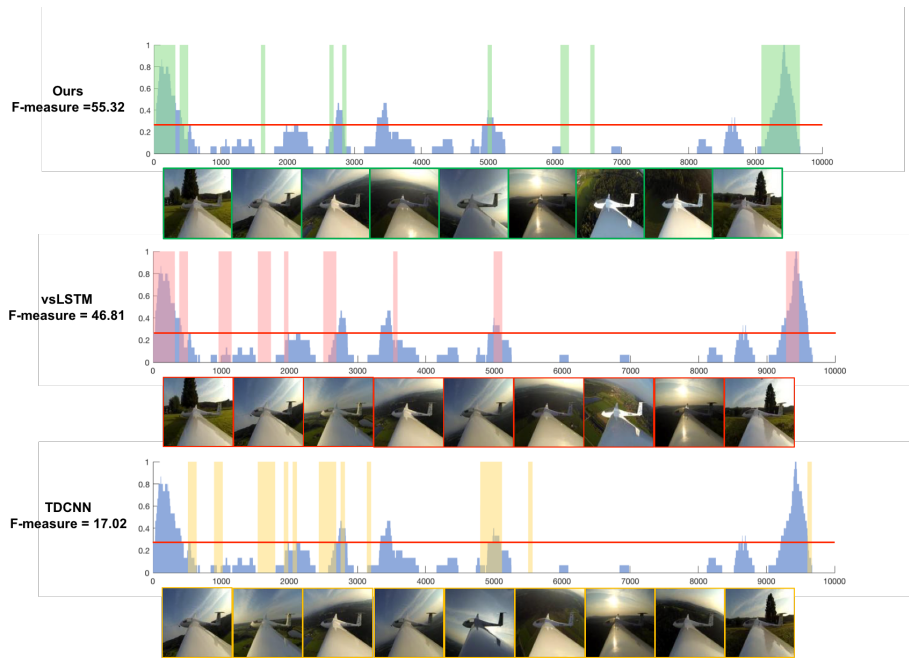

Fig. F: Example summarization results of video "Uncut Evening Flight" from SumMe. Note that our method captures moments like take-off, landing, or particular sunset scenes in the summarization output.

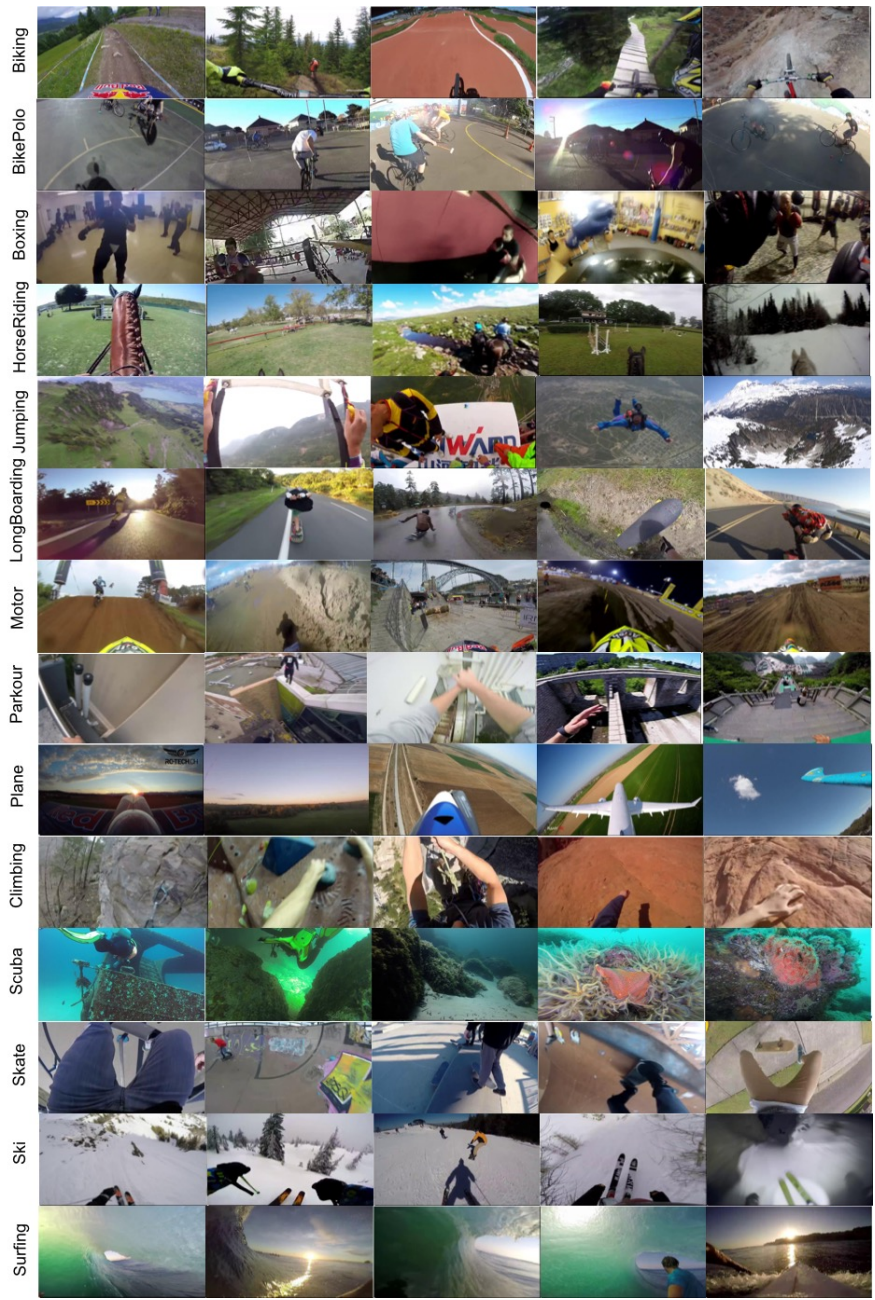

Fig. G: Thumbnails of videos in FPVSum dataset, which consists of 98 first-person videos in 14 categories captured by wearable devices from YouTube.